\documentclass[10pt,twocolumn,letterpaper]{article}

\usepackage[pagenumbers]{wacv} % To force page numbers, e.g. for an arXiv version

\usepackage{graphicx}
\usepackage{booktabs}
\usepackage{mathtools}
\usepackage{multirow}
\usepackage{amsmath,amsfonts,amssymb,bbm}
\usepackage{wrapfig,lipsum}
\usepackage{tablefootnote}
\usepackage{mathabx}
\usepackage{algorithm}
\usepackage{algorithmic}
\usepackage{colortbl}
\usepackage{tabularx}
\usepackage{caption}
\definecolor{mygray}{gray}{0.9}

\DeclareMathOperator*{\argmin}{arg\,min}

\usepackage[pagebackref,breaklinks,colorlinks]{hyperref}

\usepackage[capitalize]{cleveref}
\crefname{section}{Sec.}{Secs.}
\Crefname{section}{Section}{Sections}
\Crefname{table}{Table}{Tables}
\crefname{table}{Tab.}{Tabs.}

\def\wacvPaperID{743} % *** Enter the WACV Paper ID here
\def\confName{WACV}
\def\confYear{2025}

\begin{document}

\title{Background-aware Moment Detection for Video Moment Retrieval}

\author{
    Minjoon Jung\textsuperscript{\rm 1}~~~
    Youwon Jang\textsuperscript{\rm 2}~~~
    Seongho Choi\textsuperscript{\rm 2}~~~ 
    Joochan Kim\textsuperscript{\rm 2}~~~ \\
    \vspace{1mm}
    Jin-Hwa Kim\textsuperscript{\rm 3,4 {$\ast$}}~~~ 
    Byoung-Tak Zhang\textsuperscript{\rm 1,2,3}\thanks{\,\,Corresponding authors.} \\
    \vspace{1mm}
   \textsuperscript{1}IPAI \& \textsuperscript{2}CSE \& \textsuperscript{3}AIIS, Seoul National University~~\textsuperscript{4}NAVER AI Lab \\
   \texttt{\{mjjung,\,ywjang,\,shchoi,\,jckim,\,btzhang\}@bi.snu.ac.kr}\\
   \texttt{j1nhwa.kim@navercorp.com}
}
\maketitle

\begin{abstract}
Video moment retrieval (VMR) identifies a specific moment in an untrimmed video for a given natural language query. This task is prone to suffer the weak alignment problem innate in video datasets. Due to the ambiguity, a query does not fully cover the relevant details of the corresponding moment, or the moment may contain misaligned and irrelevant frames, potentially limiting further performance gains. To tackle this problem, we propose a background-aware moment detection transformer (BM-DETR). Our model adopts a contrastive approach, carefully utilizing the negative queries matched to other moments in the video. Specifically, our model learns to predict the target moment from the joint probability of each frame given the positive query and the complement of negative queries. This leads to effective use of the surrounding background, improving moment sensitivity and enhancing overall alignments in videos. Extensive experiments on four benchmarks demonstrate the effectiveness of our approach. Our code is available at: \url{https://github.com/minjoong507/BM-DETR}
\end{abstract}
\vspace{-2mm}    
\section{Introduction}
Video moment retrieval (VMR) \cite{gao2017tall} retrieves the target moment in an untrimmed video corresponding to a natural language query.
A successful VMR model requires a comprehensive understanding of videos, language queries, and correlations to predict relevant moments precisely.
In contrast to traditional action localization tasks \cite{yeung2016end, shou2016temporal} that predict a fixed set of actions like ``throwing'' or ``jumping,'' VMR is a more difficult task requiring joint comprehension of semantic meanings in video and language. 

\begin{figure*}[t]
    \begin{minipage}[b]{1.0\linewidth}
        \centering
        \includegraphics[width=0.99\linewidth]{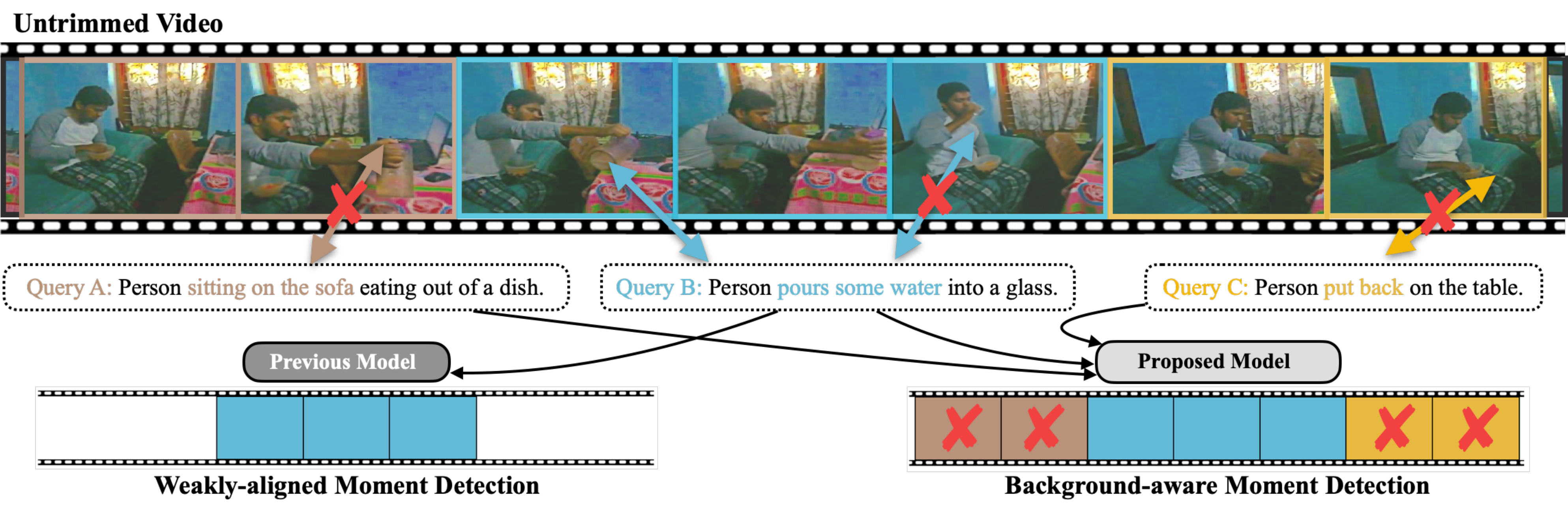}
    \end{minipage}
    \caption{
    \textit{Top:} An example of the weak alignment problem.
    \textit{Bottom:} Comparison between traditional (left) and proposed (right) methods. 
    }
    \label{fig:observation}
    \vspace{-3mm}
\end{figure*}

A video is typically composed of short video clips, where query sentences describe each clip.
However, query sentences are often ambiguous as to whether they fully express the events occurring within the matching moment, and boundary annotations might include frames unrelated to the query sentences \cite{zhou2021embracing, huang2022uncert}.
As shown in Figure \ref{fig:observation} (\textit{top}), the moment prediction can be imprecise and weakly aligned with annotations.
For instance, the query ``Person pours some water into a glass'' does not describe an event for ``drink water'', but the boundary annotation includes it.
Furthermore, queries like ``Person sitting on the sofa eating out of a dish'' may confuse the model, as the actions ``sitting'' and the object ``sofa'' overlap throughout the video. These ambiguities and weak alignments pose challenges for models in predicting specific video moments precisely.
While one might consider improving dataset quality by collecting meticulously curated video-query pairs, such approaches are often prohibitively expensive and impractical. 
In response, we aim to train an effective model that is robust even in the weak alignment problem.

Traditional VMR methods~\cite{zhang20202dtan, zhang2020span, mun2020local, zeng2020drn, gao2021fast, liu2021context, zhang2021parallel} take a single query as input to predict the moment.
However, solely relying on a single query may learn only local-level alignment and hinder achieving successful VMR due to the weak alignment problem. 
Whereas, contrastive learning-based methods~\cite{ding2021support, nan2021interventional, wang2022negative, li2023g2l} learn the query and the ground-truth moment features close to each other while keeping others apart \cite{oord2018infonce}.
Nevertheless, due to semantic overlap and sparse annotation dilemma~\cite{zeng2020drn} in videos, \cite{li2023g2l} claimed that adopting vanilla contrastive learning into VMR is suboptimal.
Negative queries from random videos used to have semantic overlap, making them false negatives, while negative moments also likely are false negatives due to the sparse annotation.
This often leads to inaccurate estimation of marginal distribution used in contrastive methods of InfoNCE~\cite{oord2018infonce}.
To overcome this, G2L~\cite{li2023g2l} employs geodesic distance to measure semantic relevance between video moments correctly, but they still need to sample a large number of negative moments, resulting in high computational costs to approximate the true partition faithfully and achieve sophisticated alignment.

In this paper, we propose a novel Background-aware Moment DEtection TRansformer (BM-DETR) based on the transformer encoder-decoder architecture~\cite{vaswani2017attention}. 
Our encoder utilizes contexts outside of the target moments (\ie, negative queries) along with the positive query.
We use the probabilistic frame-query matcher to calculate the joint probability of each frame given a positive query and the complement of negative queries, resulting in frame attention scores for enhancing multimodal representations. 
By considering the relative relationships between queries within the video, the model learns how to best identify and focus on the relevant visual features of the target moment, improving \textit{moment sensitivity}, or \textit{true positive rate}.
Then, we utilize cross-modal discrimination between other video-query pairs to learn semantic alignment.
Finally, our decoder generates predictions from multimodal features using learnable spans.
In addition, we develop the temporal shifting as an auxiliary to improve the model's robustness.

In contrast to previous approaches, which relied on a single query with complex multimodal reasoning or mining a multitude of negative moments with high cost, our model can attend to the target moment and be aware of the contextual meanings throughout the video, as shown in Figure~\ref{fig:observation} (\textit{bottom}).
Moreover, our method is simple and more efficient than previous contrastive methods by eliminating dense moment features and reducing redundant computations.
To show our model's effectiveness, we conduct experiments across four public VMR benchmarks: Charades-STA~\cite{gao2017tall}, ActivityNet-Captions~\cite{krishna2017dense}, TACoS~\cite{regneri2013tacos}, and QVHighlights~\cite{lei2021detecting}.
In addition, we provide out-of-distribution (OOD) testing and empirical qualitative analyses to further validate our findings.
To sum up, our contributions can be summarized as follows:
\begin{itemize}
    \item We propose a Background-aware Moment DEtection TRansformer (BM-DETR) to enable robust moment detection, addressing VMR challenges.
    \item BM-DETR shows significant performance improvements across four public benchmarks and also demonstrates its robustness in out-of-distribution testing.
    \item We conduct comprehensive ablation and qualitative analyses of our proposed approaches, providing deeper insights into their effectiveness.
\end{itemize}

\section{Related Work}
\noindent\textbf{Video moment retrieval.}
Video moment retrieval (VMR) aims to retrieve the target moment in a video based on a natural language sentence.
Existing approaches are mainly classified into proposal-based methods and proposal-free methods.
The proposal-based methods \cite{gao2017tall, xu2019multilevel, anne2017localizing, chen2018temporally, zhang2019cross, zhang20202dtan, gao2021fast, wang2022negative} sample candidate moments from the video and select the most similar moment to the given query.
In contrast, proposal-free methods \cite{yuan2019find, he2019read, rodriguez2020proposal, chen2020rethinking, zhang2020span, mun2020local, zeng2020drn, liu2021context, zhang2021parallel} regress target moments from video and language features without generating candidate moments.
Recently, several studies \cite{lei2021detecting, cao2021pursuit, liu2022umt, moon2023query, lin2023univtg, li2023momentdiff} have proposed DETR-based~\cite{zhu2020detr} methods.
QD-DETR~\cite{moon2023query} introduces a query-dependent to fully exploit the information from a user query.
UniVTG~\cite{lin2023univtg} focuses on developing a unified model that generalizes across multiple tasks, such as highlight detection and summarization tasks.
While previous DETR-based works focus on jointly solving localization tasks, we introduce a robust model in the weak alignment for VMR.

\noindent\textbf{Video-text alignment problem.}
Labeling videos is expensive and cumbersome, making it difficult to build high-quality and scalable video datasets. 
This often leads to alignment problems, which have been observed in previous studies~\cite{miech2020end, ko2022video, han2022temporal} as a crucial bottleneck of video understanding.
VMR is also sensitive to these issues since it requires accurate temporal moment locations, and several studies~\cite{zhou2021embracing, nan2021interventional, ding2021support, huang2022uncert, li2023g2l} are related to these problems.
To mitigate uncertainties in annotations, DeNet~\cite{zhou2021embracing} augments the phrases (e.g., verb) in language queries to improve semantic diversity, and EMB~\cite{huang2022uncert} proposes a sophisticated moment matching method.
However, since they still follow traditional VMR methods, the weak alignment problem may hinder them from achieving successful VMR.
In contrastive learning approaches, IVG-DCL~\cite{nan2021interventional} introduces the causality-based model to diminish spurious correlations between videos and queries.
G2L~\cite{li2023g2l} proposes a geodesic-guided contrastive learning scheme to reflect semantic relevance between video moments.
However, these methods still require mining numerous negative samples with high costs to optimize InfoNCE loss.
BM-DETR mitigates these dependencies and yields notable improvements in performance and efficiency.
\begin{figure*}[t]
    \begin{minipage}[b]{1.0\linewidth}
        \centering
        \includegraphics[width=0.99\linewidth]{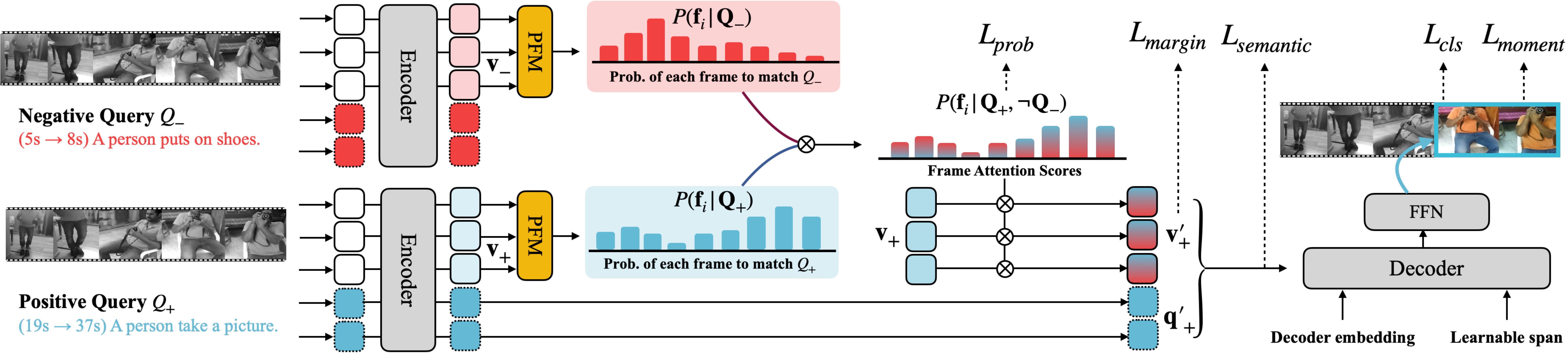}
    \end{minipage}
    \caption{
    An overview of our BM-DETR framework. 
    First, our encoder extracts multimodal features from given inputs.
    Then, we obtain frame attention scores for updating multimodal features.
    Finally, to complete VMR, our decoder predicts the target moment, and we calculate the losses from the prediction and ground-truth moment.
    }
    \label{fig:model}
    \vspace{-5mm}
\end{figure*}

\section{Method}
We give an overall architecture of the BM-DETR in Figure~\ref{fig:model}.
In this section, we first briefly review our task. 
Then, we discuss the main idea of background-aware moment detection and describe the details of our model architecture for employing it.
In Section~\ref{sec:ts}, we introduce the temporal shifting method that encourages the model's time-equivariant predictions. 
Finally, we describe how our model generates the predictions and provide details of loss.

\subsection{Video Moment Retrieval}
Given an untrimmed video $V$ and language query $Q$, we represent the video as $V=\{f_i\}_{i=1}^{L_v}$ where $f_i$ denotes the $i$-th frame. 
Likewise, the language query is denoted as $Q=\{w_i\}_{i=1}^{L_w}$ where $w_i$ denotes the $i$-th word.
$L_{v}$ and $L_w$ indicate the overall count of frames and words, respectively.
We aim to localize the target moment $m = (t_s, t_e)$ in $V$ from $Q$, where $t_s$ and $t_e$ represent the start and end times of the target moment, respectively.

\subsection{Background-aware Moment Detection}
\label{sec:background-aware moment detection}
As mentioned earlier, a single query may not be sufficient to disambiguate the corresponding moment due to the weak alignment problem in videos.
That said, predicting the target moment in $V$ based solely on information from $Q$ is less informative and ineffective, where the term “information” refers to the knowledge or cues used for accurate predictions of the target moment in $V$.
Hence, we propose an alternative to resolve this problem inspired by \textit{importance sampling}~\cite{tokdar2010importance}.
Similar to the contrastive learning~\cite{oord2018infonce}, a specific query $Q_{+}$ is designated as the target (positive), while we randomly sampled a negative query $Q_{-}$ for each training.
Our main idea is based on two guiding principles:

\textbf{Principle 1.} \textit{Queries from the same video $V$ allow for disambiguation of the target query $Q_+$, as they have implicit contextual and temporal relationships with the corresponding video moments.}

\textbf{Principle 2.} \textit{To avoid spurious correlations, we differentiate between negative query $Q_{-}$ and target query $Q_+$ based on their temporal locations and semantic similarity.}

\noindent We use $Q_{-}$ that has less intersection over union (IoU) with $Q_+$ than a certain threshold (\eg~0.5).
Additionally, we remove $Q_{-}$ that have high semantic similarity with $Q_{+}$ using SentenceBERT~\cite{reimers2019sbert} to reduce semantic overlap further.
% We will address the impact of these processes in Section~\ref{sec:ablations}.

Let $P(f_{i} \mid Q_{+})$ and $P(f_{i} \mid Q_{-})$ to be the likelihood of $i$-th frame to match the positive and negative queries, respectively.
We assume these likelihoods are independent, as their corresponding moments are at different temporal locations, and their semantic meanings are dissimilar.
Our model predicts the target moment by the joint probability of each frame, and the probability can be represented as:
\begin{equation}
    \label{eq:joint probability}
    % \tiny
    P(f_{i} \mid Q_{+}, \neg Q_{-}) \coloneqq P(f_{i} \mid Q_{+}) \cdot (1 - P(f_{i} \mid Q_{-})).
\end{equation}
Considering $P(f_i | Q_-)$, our model can focus on relatively more important meanings included in the target query, improving \textit{moment sensitivity}.
As we will see in the later experiments, being aware of contexts surrounding the target moment is more informative for the model's prediction, improving further accuracy.

\subsection{Architecture}
\label{sec:arch}
\subsubsection{Encoder.}
Our encoder aims to catch the multimodal interaction between video $V$ and query $Q$.
Initially, the pre-trained feature extractor (\eg, CLIP \cite{radford2021clip}) is employed to convert each input into multi-dimensional features and normalize them.
We utilize two projection layers to convert input features into the same hidden dimension $d$. Each projection layer consists of several MLPs. 
Then, we obtain video representations as $\mathbf{V} \in \mathbb{R}^{L_v \times d}$ and query representation as $\mathbf{Q} \in \mathbb{R}^{L_w \times d}$.
Note that there are two query representations $\mathbf{Q}_{+}$ and $\mathbf{Q}_{-}$ for positive and negative queries, respectively.
We direct them to the multimodal encoder $E(\cdot)$, a stack of transformer encoder layers denoted as: 
\begin{equation}
    E(\mathbf{V},\mathbf{Q}) = E(\textit{PE}(\mathbf{V}) \mathbin\Vert \mathbf{Q}),
\end{equation}
where \textit{PE} means the positional encoding function \cite{vaswani2017attention}, $\mathbin\Vert$ indicates the concatenation on the feature dimension. 
Finally, we obtain multimodal features $X_{+}$ and $X_{-}$ represented as:
\begin{equation}
    X_{+} = E(\mathbf{V}, \mathbf{Q}_{+}),~~X_{-} = E(\mathbf{V}, \mathbf{Q}_{-}),
\end{equation}
where $X_{+},~X_{-} \in \mathbb{R}^{L \times d}$, and we denote the length of concatenated features as $L= L_v + L_w$.

\subsubsection{Implementing the Background-aware Moment Detection.}
\label{sec:Implementing BMD}
Let us redefine the frame parts of the multimodal features $X_{+}$ and $X_{-}$ as $\mathbf{v}_{+} = \{\mathbf{f}_{i}^{+}\}_{i=1}^{L_v}$ and $\mathbf{v}_{-} = \{\mathbf{f}_{i}^{-}\}_{i=1}^{L_v}$, respectively.
We compute the likelihood of each frame to match the positive and negative queries, denoted as $P(\mathbf{f}_i \mid \mathbf{Q}_+)$ and $P(\mathbf{f}_i \mid \mathbf{Q}_-)$, respectively.
These probabilities can be obtained through the \textbf{P}robabilistic \textbf{F}rame-Query \textbf{M}atcher (PFM) defined as:
\begin{equation}
    % \small
    P(\mathbf{f}_i \mid \mathbf{Q}_{+}) = \text{PFM}(\mathbf{f}^{+}_i),~~P(\mathbf{f}_i \mid \mathbf{Q}_{-}) = \text{PFM}(\mathbf{f}^{-}_i).
\end{equation}
PFM consists of two linear layers followed by tanh and sigmoid ($\sigma$) functions defined as:
\begin{equation}
    \text{PFM}(\mathbf{f}_{i})= \sigma(\textit{tanh}(\mathbf{f}_{i}\textbf{W}_1)\textbf{W}_2),
\end{equation}
where $\mathbf{W}_1 \in \mathbb{R}^{d \times \frac{d}{2}}$ and $\mathbf{W}_2 \in \mathbb{R}^{\frac{d}{2} \times 1}$ are learnable matrices.
The joint probability of $i$-th frame $\mathbf{p}_i$ can be calculated according to Equation~\ref{eq:joint probability} as follows: 
\begin{equation}
    \label{eq: prob}
    P(\mathbf{f}_{i} \mid \mathbf{Q}_{+}, \neg \mathbf{Q}_{-}) = P(\mathbf{f}_{i} \mid \mathbf{Q}_{+}) \cdot (1 - P(\mathbf{f}_{i} \mid \mathbf{Q}_{-})).
\end{equation}
After that, the softmax function is applied to obtain the frame attention scores $\mathbf{o}$:
\begin{equation}
    \label{eq: att}
    \mathbf{o} = \text{Softmax}(\mathbf{p}_1, \mathbf{p}_2,...,\mathbf{p}_{L_v}).
\end{equation}
Finally, we leverage $\mathbf{o}$ to enhance the positive frame features $\mathbf{v}_{+}$ in $X_{+}$ to $\mathbf{v'}_+$.
This can be formulated as follows: 
\begin{equation}
    \mathbf{v'}_+ = \mathbf{o} \otimes \mathbf{v}_+,
\end{equation}
where $\otimes$ is an element-wise product.
We denote the updated multimodal features as $X_{+}'$ and send it to the decoder to predict the target moment.
Note that we only use positive queries to update multimodal features if negative queries are unavailable or during inference. 
More precisely, we substitute Equation ~\ref{eq: prob} with $P(\mathbf{f}_{i} \mid \mathbf{Q}_{+}, \neg \mathbf{Q}_{-}) = P(\mathbf{f}_{i}\mid \mathbf{Q}_{+})$.

\subsubsection{Fine-Grained Semantic Alignment.}
\label{sec: semantic alignment}
After encoding multimodal features, we focus on improving semantic alignment between video-query pairs.
Let the visual and textual representations from the multimodal features $X'_+$ are $\mathbf{v'} \in \mathbb{R}^{L_v \times d}$ and $\mathbf{q'} \in \mathbb{R}^{L_v \times d}$, respectively.
We first adopt an attentive pooling to extract global context of each representation as:
% To further enhance the effectiveness of our Background-aware Moment Detection, we adopt an attentive pooling to extract global context of each representation as:
\begin{align}
\mathbf{\hat{v}} &= \sum^{L_v}_{n=1}\mathbf{a}^{v}_{i}\mathbf{v'}_i,~~ \mathbf{a}^{v} = \text{Softmax}(\mathbf{v'}\textbf{W}_v), \\
\mathbf{\hat{q}} &= \sum^{L_v}_{n=1}\mathbf{a}^{q}_{i}\mathbf{q'}_i,~~ \mathbf{a}^{q} = \text{Softmax}(\mathbf{q'}\textbf{W}_q),
\end{align}
where $\textbf{W}_v \in \mathbb{R}^{d \times 1}$ and $\textbf{W}_q \in \mathbb{R}^{d \times 1}$ are learnable matrices.
Then we can compute the semantic alignment score as:
\begin{equation}
    \label{eq:semantic alignment score}
    S(\mathbf{\hat{v}}, \mathbf{\hat{q}}) = \frac{\mathbf{\hat{v}}^T \cdot \mathbf{\hat{q}}}{\mathbin\Vert \mathbf{\hat{v}}\mathbin\Vert_2 \mathbin\Vert\mathbf{\hat{q}}\mathbin\Vert_2},
\end{equation}
where $\mathbin\Vert \cdot \mathbin\Vert_{2}$ represents the L2-norm of a vector.
Finally, we utilize semantic alignment scores obtained from video-query pairs in our loss term (See Section~\ref{sec: learning objectives}).

\subsubsection{Decoder.}
\label{sec:Decoder}
We introduce the learnable spans to effectively use multimodal features in the moment prediction process, inspired by DAB-DETR \cite{liu2022dab}.
Instead of naively initializing the queries with learnable embeddings as in previous work~\cite{lei2021detecting}, we utilize learnable spans $S=\{S_m\}_{m=1}^M$, using moment locations as queries directly.
Each learnable span is represented as $S_m = (c_m, w_m)$, where $c_m$ and $w_m$ refer to the center and width of the corresponding span. 
We utilize positional encoding and MLP layers to generate positional query $P_m$ as:
\begin{equation}
    P_m = \text{MLP}(\textit{PE}(S_m)) = \text{MLP}(\textit{PE}(c_m) \mathbin\Vert \textit{PE}(w_m)),
\end{equation}
where \textit{PE} means fixed positional encoding to generate sinusoidal embeddings from the learnable span.
Two key modules in our decoder are self-attention and cross-attention.
In the self-attention module, the queries and keys additionally take $P_m$ as:
\begin{equation}
Q_m = D_m + P_m, \ \ K_m = D_m + P_m, \ \ V_m = D_m, \\
\end{equation}
where $D_m$ is the decoder embedding, which is initialized as 0.
Each component in the cross-attention module can be represented as:
\begin{equation}
Q_m = (D_m \mathbin\Vert \textit{PE}(S_m) \otimes \sigma(\text{MLP}(D_m))),
\end{equation}
\begin{equation}
K_m = (X_{+}' \mathbin\Vert \textit{PE}(X_{+}')), 
\end{equation}
\begin{equation}
    V_m = X_{+}'.
\end{equation}

The learnable spans are updated layer-by-layer, and please refer to~\cite{liu2022dab} for more details.

\subsection{Temporal Shifting}
\label{sec:ts}
A couple of studies \cite{xu2021boundary, zhang2021towards, hao2022can, zhang2022unsupervised} demonstrated that temporal augmentation techniques are effective for localization tasks.
Inspired by this, we design the temporal shifting method that randomly moves the ground-truth moment to a new temporal location. 
This requires our model to accurately predict based on the repositioned ground truth and background moments, allowing it to make time-equivariant predictions.
However, we acknowledge that this technique may disrupt long-term temporal semantic information in videos.
To address this issue, we empirically apply the temporal shifting method to videos with short durations (\ie,~$|V|$ $<$ 60 s).
Further details with a visual example are provided in the Supplementary.

\subsection{Learning Objectives}
\label{sec: learning objectives}
\noindent\textbf{Predictions.}
Based on the decoder outputs, we apply MLP layers to generate a set of $M$ predictions denoted as $\hat{y} = \{\hat{y}_i\}_{i=1}^{M}$.
Each prediction $\hat{y}_i$ contains two components: 1) the class label $\hat{c}_i$ to indicate whether the predicted moment is the ground-truth moment or not, and 2) temporal moment location $\hat{m}_i = (\hat{t}_{s}^i, \hat{t}_{e}^{i})$.
Following the previous work \cite{lei2021detecting}, we find the optimal assignment $i$ between the ground-truth $y$ and the predictions $\hat{y}_i$ using Hungarian algorithm based on the matching cost $\mathcal{C}_{\rm  match}$ as:
\begin{equation}
    \mathcal{C}_{\rm  match}(y, \hat{y}_{i}) = -p(\hat{c}_i) + \mathcal{L}_{\rm  moment}(m, \hat{m}_i),
\end{equation}
\begin{equation}
  i = \argmin_{i \in N}\mathcal{C}_{\rm  match}(y, \hat{y}_{i}).
\end{equation}

\noindent\textbf{Moment localization loss.} 
The moment localization loss contains two losses: 1) $\rm L1$ loss and 2) a generalized IoU loss \cite{rezatofighi2019generalized}.
This loss is designed to calculate the accuracy of a prediction by comparing it to the ground-truth moment.
\begin{equation}
    % \small
    \mathcal{L}_{\rm moment}(m, \hat{m}_i) = \lambda_{\rm L1} \mid\mid m - \hat{m}_i \mid\mid + \lambda_{\rm iou}  \mathcal{L}_{\rm iou}(m, \hat{m}_i),
\end{equation}
where $\lambda_{\rm L1}$ and $\lambda_{\rm iou}$ are the coefficients to adjust weights.

\noindent\textbf{Frame margin loss.} 
The margin loss encourages frames within the ground-truth moment to have high scores via hinge loss.
We use a linear layer to predict the scores of the frame features $\mathbf{f_{\rm fore}}$ and $\mathbf{f_{\rm back}}$ within $\mathbf{v}_{+}'$.
Note that $\mathbf{f_{\rm fore}}$ is located within the ground-truth moment, and $\mathbf{f_{\rm back}}$ is not.
The loss can be formulated as follows: 
\begin{equation}
  \label{eq:margin}
  \mathcal{L}_{\rm margin} = \mathrm{max}(0, \Delta + \mathbf{f}_{\rm back}\mathbf{W} - \mathbf{f}_{\rm fore}\mathbf{W}),
\end{equation}
where $\mathbf{W} \in \mathbb{R}^{d \times 1}$, and we set the margin $\Delta$ as 0.2.

\noindent\textbf{Frame probability loss.}
We encourage frames within the target moment to have a high probability.
Let $\mathcal{P}$ and $\mathcal{N}$ be the sets of frame indices $\mathbf{f_{\rm fore}}$ and $\mathbf{f_{\rm back}}$.
Then we calculate the loss from the joint probability of frames $\mathbf{p}=\{\mathbf{p}_{i}\}_{i}^{L_v}$ (in Equation~\ref{eq: prob}) as follows:
\begin{equation}
    \mathcal{L}_{\rm prob} = 1 - \frac{1}{|\mathcal{P}|}\sum_{i \in \mathcal{P}}\mathbf{p}_{i} + \frac{1}{|\mathcal{N}|}\sum_{j \in \mathcal{N}}\mathbf{p}_{j}.
\end{equation}

\noindent\textbf{Semantic alignment loss.} 
We denote video-query pairs within a batch $B$ as $\{V_i, Q_i\}_{i=1}^{N}$. 
We obtain multimodal features $\{\mathbf{\hat{v}}_i, \mathbf{\hat{q}}_i \}_{i=1}^{N}$ and compare the semantic alignment scores (in Equation~\ref{eq:semantic alignment score}) between positive and negative video-query pairs.
The loss can be formulated as follows:
\begin{equation}
  \mathcal{L}_{\rm semantic} = - \frac{1}{|N|} \sum_{i=1}^{N}\mathrm{log}\frac{\mathrm{exp}(S(\mathbf{\hat{v}}_i, \mathbf{\hat{q}} _i) / \tau)}{\sum_{j \in \mathcal{N}}\mathrm{exp}(S(\mathbf{\hat{v}}_i, \mathbf{\hat{q}}_j) / \tau)},
\end{equation}
where $\tau$ is a temperature parameter and set as 0.07, and $\mathcal{N}$ means negative pairs $\{(\mathbf{\hat{v}}_i, \mathbf{\hat{q}}_j): i \neq j\}$.

\noindent\textbf{Overall loss.} 
The overall loss is a linear combination of individual losses.
Additionally, we use the class loss (\ie,~$\mathcal{L}_{\rm cls}$), which is the cross-entropy function computed by $\hat{c}_i$ that classifies whether the predicted moment is the ground-truth moment.
Also, we set the hyper-parameters for each loss term to adjust the scale of loss.
\section{Experiments}
\begin{table}[t!]
    \small
    \centering
    \resizebox{\linewidth}{!}{
        \begin{tabular}{lccccc}
        \toprule
        \multirow{2}{*}{Dataset} & \multirow{2}{*}{Domain} & \multirow{2}{*}{\#Videos} & \multirow{2}{*}{\#Queries} & Avg (sec) & Avg \\  
        & & & & Moment/Video & Query \\ \midrule
        CharadesSTA & Activity & 6.7K & 16.1K & 8.1 / 30.6 & 7.2 \\
        Anet-Cap & Activity & 15K & 72K & 36.2 / 117.6 & 14.8 \\
        TACoS & Cooking & 127 & 18K & 5.4 / 287.1 & 10 \\
        QVHighlights & Vlog / News & 10.2K & 10.3K & 24.6 / 150 & 11.3 \\
        \bottomrule
        \end{tabular}
    }
    \vspace{-2mm}
    \caption{
    Statistics of VMR datasets.
    Avg Moment/Video denotes an average length of moment/video in seconds.
    Avg Query means an average number of words in query sentences.
    }
    \label{tbl:datasets}
    \vspace{-5mm}
\end{table}
\subsection{Experiment Setup}
\noindent\textbf{Datasets.} We experiment on four VMR datasets: Charades-STA \cite{gao2017tall}, Activitynet-Captions~\cite{krishna2017dense}, TACoS \cite{regneri2013tacos}, and QVHighlights \cite{lei2021detecting} test splits and provide the statistics of datasets in Table~\ref{tbl:datasets}.

\begin{table}[t!]
    \begin{center}
  \begin{tabular}{l|l|cc}
  \toprule
    \multirow{2}{*}{Method} & \multirow{2}{*}{Video Feat} & \multicolumn{2}{c}{\textbf{Charades-STA}} \\ 
    \cline{3-4} & & IoU=0.5 & IoU=0.7 \\ \midrule
    % For C3D,
    2D-TAN \cite{zhang20202dtan} & \multirow{7}{*}{C3D} & 39.70 & 27.10   \\
    DRN \cite{zeng2020drn} & & 45.40	& 26.40 \\
    VSLNet \cite{zhang2020span} & & 47.31 & 30.19 \\
    CBLN \cite{liu2021context} & & 47.94 & 28.22 \\
    IVG-DCL \cite{nan2021interventional} & & 50.24  & \underline{32.88} \\
    MomentDiff~\cite{li2023momentdiff} & & \underline{53.79} & 30.18 \\
    \bf BM-DETR (ours) & & \bf 54.42 & \bf 33.84 \\ \midrule
    % For VGG,
    2D-TAN \cite{zhang20202dtan} & \multirow{10}{*}{VGG} & 41.34 & 23.91   \\
    DRN \cite{zeng2020drn} & & 42.90 & 23.68 \\
    CBLN \cite{liu2021context} & & 43.67 & 24.44 \\
    FVMR \cite{gao2021fast} & & 42.36 & 24.14 \\
    SSCS \cite{ding2021support} & & 43.15 & 25.54 \\ 
    % UMT* \cite{liu2022umt} & & 48.31 & 29.25 \\
    MMN \cite{wang2022negative} & & 47.31 & 27.28  \\
    QD-DETR \cite{moon2023query} & & \underline{52.77} & \underline{31.13} \\
    G2L \cite{li2023g2l} & & 47.91 & 28.42 \\ 
    MomentDiff~\cite{li2023momentdiff} & & 51.94 & 28.25 \\
    \bf BM-DETR (ours) & & \bf 54.22 & \bf 35.54 \\ \midrule
    % For SF+C,
    MDETR \cite{lei2021detecting}& \multirow{5}{*}{SF+C} & 53.63 & 31.37 \\
    QD-DETR \cite{moon2023query} & & 57.31 & 32.55 \\
    UniVTG \cite{lin2023univtg} & & \underline{58.01} & \underline{35.65} \\
    MomentDiff~\cite{li2023momentdiff} & & 55.57 & 32.42 \\ 
    \bf BM-DETR (ours) & & \bf 59.48 & \bf 38.33 \\ 
  \bottomrule
  \end{tabular}
  \end{center}
    \vspace{-4mm}
        \caption{Performance results on Charades-STA.} 
    \vspace{-2mm}
    \label{tbl:charades-sta}
\end{table}
\noindent\textbf{Implementation details.} 
Our model is built upon MDETR~\cite{lei2021detecting} implemented in Pytorch.
We use the same features in each dataset as used in previous models for a fair comparison: VGG \cite{simonyan2014vgg}, C3D \cite{tran2015c3d}, I3D~\cite{carreira2017i3d}, and SF+C, which is a concatenation of SlowFast \cite{feichtenhofer2019slowfast} and CLIP \cite{radford2021clip} for video features and Glove embedding~\cite{pennington2014glove}, BERT~\cite{devlin2018bert}, and CLIP~\cite{radford2021clip} for text features.
For Charades-STA, we use GloVe embeddings along with C3D and VGG, and CLIP for SF+C.
Please refer to the Supplementary for more details.

\noindent\textbf{Evaluation metric.} 
We use two metrics for comparison: 1) R@$n$, IoU=$m$, which measures the percentage of the top-n predicted moments with an IoU greater than m. 
We set $n$ as 1 and take $m$ from the threshold set $\{0.3, 0.5, 0.7\}$. 
2) We use Mean Average Precision (mAP) over IoU thresholds of 0.5 and 0.75, and calculate the average mAP (\ie, Avg.) across multiple IoU thresholds.

\begin{table}[t]
    \begin{center}
    \resizebox{\linewidth}{!}{
      \begin{tabular}{l|l|cc|cc}
      \toprule
        \multirow{3}{*}{Method} & \multirow{3}{*}{Text Feat} & \multicolumn{2}{c|}{\bf ActivityNet-Captions} & \multicolumn{2}{c}{\bf TACoS} \\
        \cline{3-6} & & \multicolumn{2}{c|}{Video Feat: C3D} & \multicolumn{2}{c}{Video Feat: C3D} \\ 
         \cline{3-6} & & IoU=0.5 & IoU=0.7 & IoU=0.3 & IoU=0.5 \\ \midrule
        % \cmidrule(lr){4-5} \cmidrule(lr){6-7} & & & R@1, IoU=0.3 & R@1, IoU=0.5 & R@1, IoU=0.3 & R@1, IoU=0.5 \\ \midrule
    2D-TAN~\cite{zhang20202dtan} & Glove & 44.51 & 26.54 & 37.29 & 25.32 \\
    VSLNet \cite{zhang2020span} & Glove & 43.22 & 26.16 & 29.61 & 24.27 \\
    DRN~\cite{zeng2020drn} & Glove & 45.45 & 24.39 & - & 23.17 \\
    CBLN \cite{liu2021context} & Glove & 48.12 & 27.60 & 38.98 & 27.65 \\
    DeNet~\cite{zhou2021embracing} & Glove & 43.79 & - & - & - \\
    IVG-DCL \cite{nan2021interventional} & Glove & 43.84 & 27.10 & 38.84 & 29.07 \\  
    SSCS \cite{ding2021support} & Glove & 46.67 & 27.56 & 41.33 & 29.56 \\ 
    GTR \cite{cao2021pursuit} & Glove & \bf 50.57 & \underline{29.11} & \underline{40.39} & \underline{30.22} \\
    \bf BM-DETR (ours) & Glove & \underline{49.62} & \bf 30.61 & \bf 49.87 & \bf 33.67 \\ \midrule
    MMN \cite{wang2022negative} & DistilBERT & 48.59 & 29.26 & 39.24 & 26.17 \\
    G2L \cite{li2023g2l} & BERT & \bf 51.68 & \bf 33.35 & \underline{42.74} & \underline{30.95} \\
    \textbf{BM-DETR (ours)} & BERT & \underline{49.98} & \underline{30.88} & \bf 50.46 & \bf 35.87 \\
      \bottomrule
      \end{tabular}
      }
    \end{center}
    \vspace{-3mm}
    \caption{
    Performance results on ActivityNet-Captions and TACoS.
    }
    % \vspace{-4mm}
    \label{tbl:anet&tacos}
\end{table}

\subsection{Comparison with the State-of-the-Art Methods}
\noindent\textbf{Baselines.} 
In this section, we compare BM-DETR with baselines, which can be divided into three categories:
1) traditional VMR methods that take a single query as input for predictions, 
2) methods based on contrastive learning (\ie,~CL-based), including IVG-DCL~\cite{nan2021interventional}, SSCS~\cite{ding2021support}, MMN~\cite{wang2022negative}, and G2L~\cite{li2023g2l},
3) methods following DETR's detection paradigm (\ie,~DETR-based), including GTR~\cite{cao2021pursuit}, MDETR~\cite{lei2021detecting}, UMT~\cite{liu2022umt}, QD-DETR~\cite{moon2023query}, UniVTG~\cite{lin2023univtg}, and MomentDiff~\cite{li2023momentdiff}.
In each table, the highest score is bolded, and the second highest score is underlined.
\begin{table}[t!]
    \small
    \centering
    \resizebox{\linewidth}{!}{
    \begin{tabular}{l|l|ccccc}
        \toprule
        \multirow{3}{*}{Method} & \multirow{3}{*}{Text Feat} & \multicolumn{5}{c}{\bf QVHighlights} \\
        % \cline{3-7} \\
        \cline{3-7} & & \multicolumn{5}{c}{Video Feat: SF+C}\\
        \cline{3-7} & & IoU=0.5 & IoU=0.7 & mAP@0.5 & mAP@0.75 & Avg. \\ \midrule
        MCN~\cite{anne2017localizing} & CLIP & 11.41 & 2.72 & 24.94 & 8.22 & 10.67 \\
        CAL~\cite{escorcia2019temporal} & CLIP & 25.49 & 11.54 & 23.40 & 7.65 & 9.89 \\
        XML~\cite{lei2020tvr} & CLIP & 41.83 & 30.35 & 44.63 & 31.73 & 32.14  \\
        XML+~\cite{lei2021detecting} & CLIP & 46.69 & 33.46 & 47.89 & 34.67 & 34.90 \\
        MDETR~\cite{lei2021detecting} & CLIP & 52.89 & 33.02 & 54.82 & 29.40 & 30.73  \\
        UMT~\cite{liu2022umt} & CLIP & 56.23 & 41.18 & 53.83 & 37.01 & 36.12 \\
        QD-DETR~\cite{moon2023query} & CLIP & \bf 62.40 & \bf 44.98 & \underline{62.52} & \underline{39.88} & \underline{39.86} \\
        UniVTG~\cite{lin2023univtg} & CLIP & 58.86 & 40.86 & 57.60 & 35.59 & 35.47 \\
        MomentDiff~\cite{li2023momentdiff} & CLIP & 57.42 & 39.66 & 54.02 & 35.73 & 35.95 \\
        \bf BM-DETR (ours) & CLIP & \underline{60.12} & \underline{43.05} & \bf 63.08 & \bf 40.18 & \bf 40.08 \\
        \bottomrule
    \end{tabular}
    }
    \vspace{-1mm}
    \caption{
    Performance results on QVHighlights.
    % ``*'' indicates the use of additional source (\ie., audio).
    }
    \vspace{-1mm}
    \label{tbl:qvhighlights}
\end{table}
\begin{table}[t!]
  \centering
  % \resizebox{\linewidth}{!}{
      \begin{tabular}{l|l|cc}
      \toprule
        \multirow{3}{*}{Method} & \multirow{3}{*}{Text Feat} & \multicolumn{2}{c}{\bf Charades-CD} \\  
        \cline{3-4} & & \multicolumn{2}{c}{Video Feat: I3D} \\
        \cline{3-4} & & IoU=0.5 & IoU=0.7 \\  \midrule 
      2D-TAN \cite{zhang20202dtan} & Glove  & 35.88  &  13.91    \\
      LG \cite{mun2020local} & Glove &  42.90  &  19.29   \\
      DRN \cite{zeng2020drn} & Glove & 31.11  &  15.17   \\
      VSLNet \cite{zhang2020span} & Glove & 34.10  &  17.87   \\
      DCM \cite{yang2021deconfounded} & Glove & 45.47  &  22.70   \\
      Shuffling \cite{hao2022can}   & Glove &  \underline{46.67}  &  \underline{27.08}    \\
      \textbf{BM-DETR (ours)} & Glove & {\bf 53.37} & {\bf 30.12} \\
      \bottomrule
      \end{tabular}
      % }
    % \vspace{-2mm}
    \caption{Performance results on Charades-CD.
  }
\vspace{-4mm}
\label{tbl:charades-CD}
\end{table}

\noindent\textbf{Comparison with traditional VMR methods.}
All results in Table~\ref{tbl:charades-sta}, and~\ref{tbl:anet&tacos} show that BM-DETR's superior performance compared to traditional VMR methods, such as 2D-TAN~\cite{zhang20202dtan} and VSLNet~\cite{zhang2020span}, across all datasets.
These results indicate that accurate predictions for these methods may be challenging due to the weak alignment problem.
In addition, they also easily suffer the bias problem in datasets, as we will discuss in the next section.
\begin{table*}[t]
  \centering
      \small
      \resizebox{\linewidth}{!}{
      \begin{tabular}{lcccccccccccc}
      \toprule
      \multirow{2}{*}{Method} & \multicolumn{3}{c}{\bf Charades-STA} & \multicolumn{3}{c}{\bf TACoS} & \multicolumn{3}{c}{\bf ActivityNet-Captions} & \multicolumn{3}{c}{\bf QVHighlights} \\
       \cmidrule(lr){2-4} \cmidrule(lr){5-7} \cmidrule(lr){8-10} \cmidrule(lr){11-13}  
       & GT $\uparrow$ & Non-GT $\downarrow$ & $\vartriangle$ $\uparrow$ 
       & GT $\uparrow$ & Non-GT $\downarrow$ & $\vartriangle$ $\uparrow$
       & GT $\uparrow$ & Non-GT $\downarrow$ & $\vartriangle$ $\uparrow$
       & GT $\uparrow$ & Non-GT $\downarrow$ & $\vartriangle$ $\uparrow$ \\ \midrule 
       Baseline & 0.42 & 0.20 & 0.22 
       & 0.56 & 0.18 & 0.38 
       & 0.52 & 0.24 & 0.28 
       & 0.67 & 0.35 & 0.32 \\
       \bf BM-DETR (ours)  & \bf 0.56 & \bf 0.13 & \bf 0.43 
       & \bf 0.60 & \bf 0.11 & \bf 0.49 
       & \bf 0.56 & \bf 0.21 & \bf 0.35 
       & \bf 0.73 & \bf 0.28 & \bf 0.45 \\ 
      \bottomrule
      \end{tabular}
      }
    % \vspace{-2mm}
    \caption{
    Evaluation of video-text alignment.
    The average of the joint probabilities of frames $\textbf{p}$ (in Equation~\ref{eq: prob}) inside and outside the ground-truth moment, denoted as GT and Non-GT, respectively.
    }
    \label{tbl:prob}
\vspace{-3mm}
\end{table*}

% \begin{table}[t]
%   \centering
%       \small
%       \resizebox{\linewidth}{!}{
%       \begin{tabular}{lcccccc}
%       \toprule
%       \multirow{2}{*}{Method} & \multicolumn{3}{c}{Charades-STA} & \multicolumn{3}{c}{TACoS} \\
%        \cmidrule(lr){2-4} \cmidrule(lr){5-7}
%        & GT $\uparrow$ & Non-GT $\downarrow$ & $\vartriangle$ $\uparrow$ 
%        & GT $\uparrow$ & Non-GT $\downarrow$ & $\vartriangle$ $\uparrow$ \\ \midrule 
%        Baseline & 0.42 & 0.20 & 0.22 
%        & 0.56 & 0.18 & 0.38 \\
%        \bf Ours  & \bf 0.56 & \bf 0.13 & \bf 0.43 
%        & \bf 0.60 & \bf 0.11 & \bf 0.49 \\ \midrule
%        & \multicolumn{3}{c}{Anet-Cap} & \multicolumn{3}{c}{QVHighlights} \\
%         & 0.52 & 0.24 & 0.28 
%        & 0.67 & 0.35 & 0.32 \\
%        & \bf 0.56 & \bf 0.21 & \bf 0.35 
%        & \bf 0.73 & \bf 0.28 & \bf 0.45 \\ 
%       \bottomrule
%       \end{tabular}
%       }
%     \vspace{2mm}
%     \caption{
%     The average of the joint probabilities of frames $\textbf{p}$ (in Equation~\ref{eq: prob}) inside and outside the ground-truth moment, denoted as GT and Non-GT, respectively.
%     }
%     \label{tbl:prob}
% % \vspace{-8mm}
% \end{table}

\noindent\textbf{Comparison with CL-based methods.}
Our model outperforms most of the contrastive learning-based methods across all datasets.
Specifically, BM-DETR outperforms the recent state-of-the-art CL-based method (\ie, G2L~\cite{li2023g2l}) by over 7.3 points in R@1, IoU=0.7 in Table~\ref{tbl:charades-sta}.
Without incorporating modules like IVG module in IVG-DCL~\cite{nan2021interventional} or captioning objectives in SSCS~\cite{ding2021support}, BM-DETR simply integrates a negative query into predictions using a lightweight module (\ie,~PFM) consisting of only two linear layers. 
This avoids the complexities in MMN~\cite{wang2022negative} and G2L~\cite{li2023g2l}, which incur high computation costs associated with negative samples to improve joint representation learning.
% Additionally, unlike MMN~\cite{wang2022negative} and G2L~\cite{li2023g2l}, which incur high computation costs associated with negative samples to improve joint representation learning, our model shows competitive performance avoiding such complexities. 
We also quantify the efficiency of BM-DETR compared with the previous methods in Table~\ref{tbl:efficiency}.

\noindent\textbf{Comparison with DETR-based methods.}
Compared with the previous DETR-based methods, our model shows competitive performance.
Notably, in Charades-STA, where annotations are often weakly aligned and noisy~\cite{nan2021interventional, ding2021support}, our model significantly outperforms previous methods in Table~\ref{tbl:charades-sta}.
This indicates that the reliance on accurate annotation of the model is reduced compared to the previous methods, demonstrating our model's robustness.

\begin{table}[t]
        \centering
        \begin{tabular}{cccccc}%{{\centering}XXXX|XX}%{cccccc}
        \toprule
        \multirow{2}{*}{BMD} & \multirow{2}{*}{FS} & \multirow{2}{*}{LS} & \multirow{2}{*}{TS} & \multicolumn{2}{c}{\bf Charades-STA} \\  
        \cmidrule(lr){5-6} & & & & {IoU=0.5} & {IoU=0.7} \\  
        \midrule
        {} & {} & {} & {} & {51.43} & {28.87} \\
        {$\checkmark$} & {} & {} & {} & {54.73} & {33.28} \\
        {} & {$\checkmark$} & {} & {} & {53.76} & {32.13} \\
        {} & {} & {$\checkmark$} & {} & {54.39} & {32.23} \\
        {} & {} & {} & {$\checkmark$} & {53.47} & {31.12} \\
        {$\checkmark$} & {$\checkmark$} & {} & {} & {55.02} & {33.64} \\
        {$\checkmark$} & {} & {$\checkmark$} & {} & {53.98} & {33.53} \\
        {$\checkmark$} & {} & {$\checkmark$} & {$\checkmark$} & {58.79} & {35.04} \\
        {$\checkmark$} & {$\checkmark$} & {$\checkmark$} & {$\checkmark$} & {\bf 59.48} & {\bf 38.33} \\
        \bottomrule
        \end{tabular}
      \vspace{-1mm}
        \caption{
        Ablations on model components.
        BMD: background-aware moment detection, FS: fine-grained semantic alignment, LS: learnable spans, and TS: temporal shifting.
        }
    \label{tbl:model_components}
    \vspace{-4mm}
\end{table}
\subsection{Out-of-Distribution Testing}
\label{sec:OOD}
Charades-STA has been widely used for VMR datasets, but there are significant bias problems \cite{otani2020uncovering, yuan2021closer} that current models tend to rely on identifying frequent patterns in the temporal moment distribution (\ie,~temporal bias) rather than real comprehension of multimodal inputs.
To enhance evaluation reliability, we conduct experiments on an out-of-distribution test split (\ie,~test-ood) in Charades-CD~\cite{yuan2021closer}.
The test-ood splits have different temporal distributions of queries from the training splits, which presents a challenge that requires the model's generalization and strong alignment capabilities.
In Table~\ref{tbl:charades-CD}, BM-DETR shows its robustness surpassing prior approaches.
While DCM~\cite{yang2021deconfounded} and Shuffling~\cite{hao2022can} are designed to address the temporal bias problem, they still follow traditional VMR methods.
These results highlight the robustness of our model and re-emphasize that relying solely on a single query might be insufficient to solve VMR challenges.

\begin{table}[t]
        \centering
        \begin{tabular}{cccccc}
      \toprule
        \multirow{2}{*}{$\mathcal{L}$} & \multirow{2}{*}{$\mathcal{L}_{\rm m}$} & \multirow{2}{*}{$\mathcal{L}_{\rm s}$} & \multirow{2}{*}{$\mathcal{L}_{\rm p}$} & \multicolumn{2}{c}{\bf Charades-STA} \\  
        \cmidrule(lr){5-6} & & & & IoU=0.5 & IoU=0.7 \\ \midrule
        {} & {$\checkmark$} & {$\checkmark$} & {$\checkmark$} & {18.36} & {5.31} \\
        {$\checkmark$} & {} & {} & {} & {29.02} & {14.63} \\
        {$\checkmark$} & {$\checkmark$} & {} & {} & {56.49} & {36.11} \\
        {$\checkmark$} & {} & {$\checkmark$} & {} & {57.42} & {36.01} \\
        {$\checkmark$} & {} & {} & {$\checkmark$} & {56.32} & {35.45} \\
        {$\checkmark$} & {$\checkmark$} & {} & {$\checkmark$} & {58.10} & {36.23} \\
        {$\checkmark$} & {} & {$\checkmark$} & {$\checkmark$} & {57.84} & {36.70} \\
        {$\checkmark$} & {$\checkmark$} & {$\checkmark$} & {} & {58.68} & {37.59} \\
        {$\checkmark$} & {$\checkmark$} & {$\checkmark$} & {$\checkmark$} & {\bf 59.48} & {\bf 38.33} \\
      \bottomrule
      \end{tabular}
      \vspace{-1mm}
        \caption{
        Ablations on losses.
        We denote each loss as 
        $\mathcal{L}$: combination of  moment localization loss and class loss,
        $\mathcal{L}_{\rm m}$: frame margin loss,
        $\mathcal{L}_{\rm s}$: semantic align loss,
        and $\mathcal{L}_{\rm p}$: frame probability loss.
        }
        \label{tbl:losses}
        \vspace{-4mm}
\end{table}
\subsection{Ablation Study}
\label{sec:ablations}
In this section, we conduct comprehensive ablation studies to provide an in-depth analysis of our approach.

\noindent\textbf{Evaluation of video-text alignment.}
To quantitatively assess the improvements in video-text alignment facilitated by our approaches,
we first set a baseline model that uses only the PFM applied to MDETR.
We then compare the average of $\textbf{p}$ (in Equation~\ref{eq: prob}) obtained by BM-DETR against the baseline model.
Table~\ref{tbl:prob} shows that BM-DETR demonstrates a clear gap of the average $\textbf{p}$ between GT and Non-GT across all datasets compared to the baseline.
These results provide clear evidence that our method enhances overall video-text alignments, successfully differentiating between ground-truth and background moments.

\noindent\textbf{Ablations on model components.} 
To validate the effectiveness of each model component, we build up several baseline models with different model components.
With the results in Table~\ref{tbl:model_components}, we confirm that all components jointly perform well and contribute to performance improvement.

\noindent\textbf{Ablations on losses.} 
In Table~\ref{tbl:losses}, we investigate the impact of each loss.
As we can see in the first row, $\mathcal{L}$ (\ie,~$\mathcal{L}_{\rm cls}$ and $\mathcal{L}_{\rm moment}$) is necessary to perform VMR as it directly guides whether the prediction matches the ground-truth.
We can see that jointly combining our losses provides significant performance gains.

\begin{figure}[t]
  \begin{center}
    \includegraphics[width=1.0\linewidth]{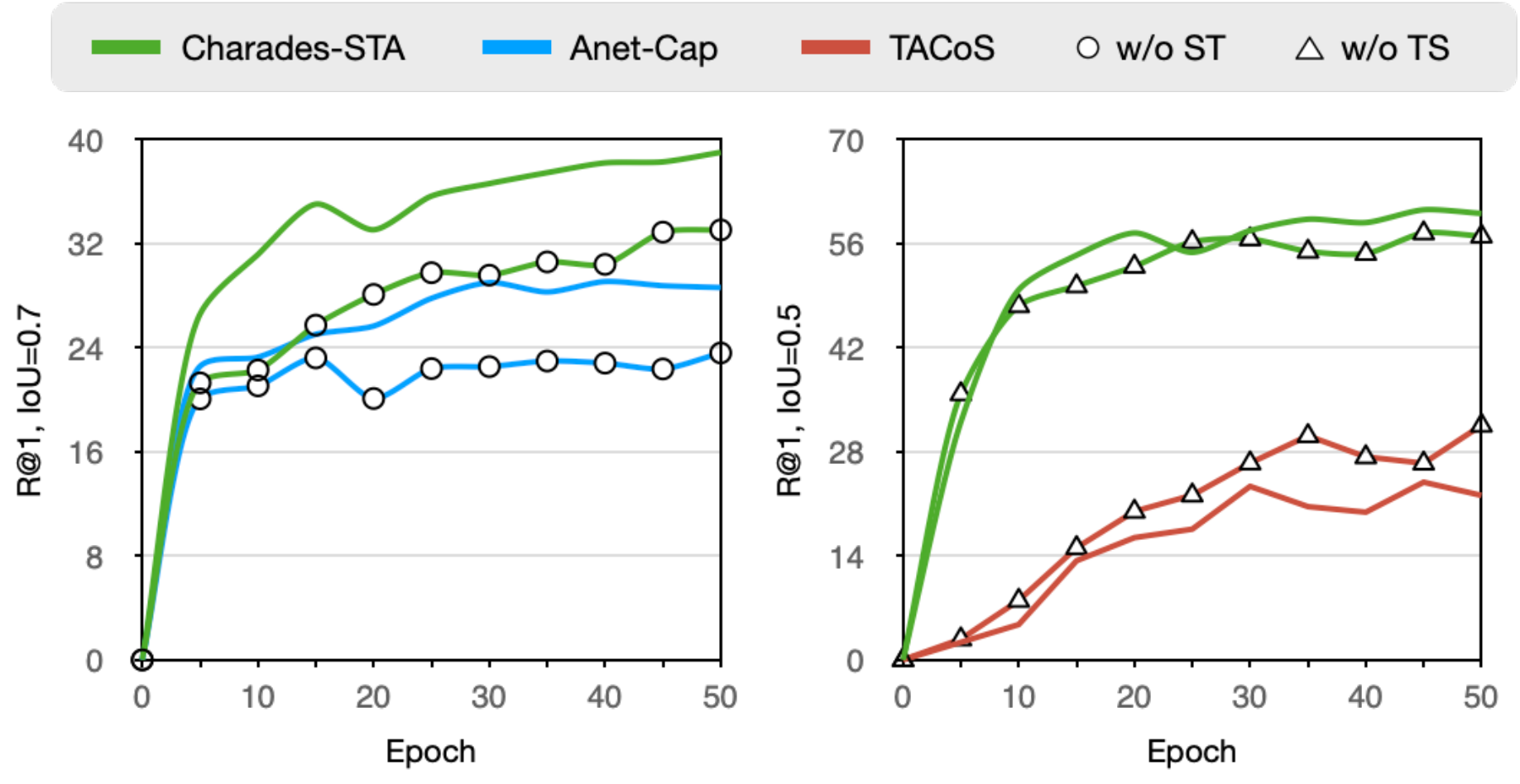}
    \end{center}
        \vspace{-3mm}
       \captionof{figure}{
          \textit{Left}: Bold circles indicate when our sampling strategy (ST) is not applied.
          \textit{Right}: Bold triangles indicate when temporal shifting (TS) is not applied.
          }
    \label{fig:st&ts}
\end{figure}

\noindent\textbf{Effect of sampling strategy.}
In Figure~\ref{fig:st&ts} (\textit{left}), we conduct experiments with and without our sampling strategy on Charades-STA and ActivityNet-Captions.
Without our sampling strategy, the model treats queries that are near the ground truth and semantically similar to target queries as negatives, resulting in significant performance degradation (around 7 points in R@1, IoU=0.7) on both datasets.
This shows the importance of sampling strategy and supports that vanilla contrastive learning may be suboptimal for VMR.
\begin{table}[t]
    \resizebox{1.0\linewidth}{!}{
      \begin{tabular}{lcccc}
      \toprule
      \multirow{2}{*}{Method} & \multirow{2}{*}{Iteration} & Total & Total & \multirow{2}{*}{\#GPU} \\  
      & & Inference & Training & \\ \midrule 
      MMN~\cite{wang2022negative} & 0.32s &  37s & 10h & 6 \\
      G2L~\cite{li2023g2l} & 0.84s & 43s & - & 8  \\
      BM-DETR (ours)  & 0.19s & 21s & 3h & 1 \\
      \bottomrule
      \end{tabular}
      }
      \vspace{-2mm}
        \captionof{table}{
          Efficiency comparison on Anet-Cap.
          The results of the other studies follow the original papers. 
        }
        \label{tbl:efficiency}
        \vspace{-3mm}
\end{table}

\noindent\textbf{Effect of temporal shifting.}
As mentioned in Section~\ref{sec:ts}, we employ temporal shifting (TS) selectively, as it can disrupt the long-term temporal context in videos.
To empirically assess this, we conduct experiments on TACoS and Charades-STA without considering video lengths.
Among the datasets used in our study, TACoS has the longest average video duration (287s), and Charades-STA has the shortest average video duration (30s).
Figure~\ref{fig:st&ts} (\textit{right}) shows improved performance for Charades-STA regardless of video lengths, whereas there is a decline for TACoS.
This indicates that longer videos are more sensitive to TS which may disrupt temporal information within the video and hinder training.
A further detailed examination of this is discussed in the Supplementary.

\noindent\textbf{Efficiency comparison.} 
In Table~\ref{tbl:efficiency}, we compare the efficiency of BM-DETR with recent CL-based methods, including MMN~\cite{wang2022negative} and G2L~\cite{li2023g2l} under the same setting.
While they require cumbersome computations between a number of negative video moments and queries, our model performs quite efficiently.

\subsection{Visualization Results}
\label{sec:visualization}
\begin{figure}[t]
  \begin{center}
    \includegraphics[width=0.99\linewidth]{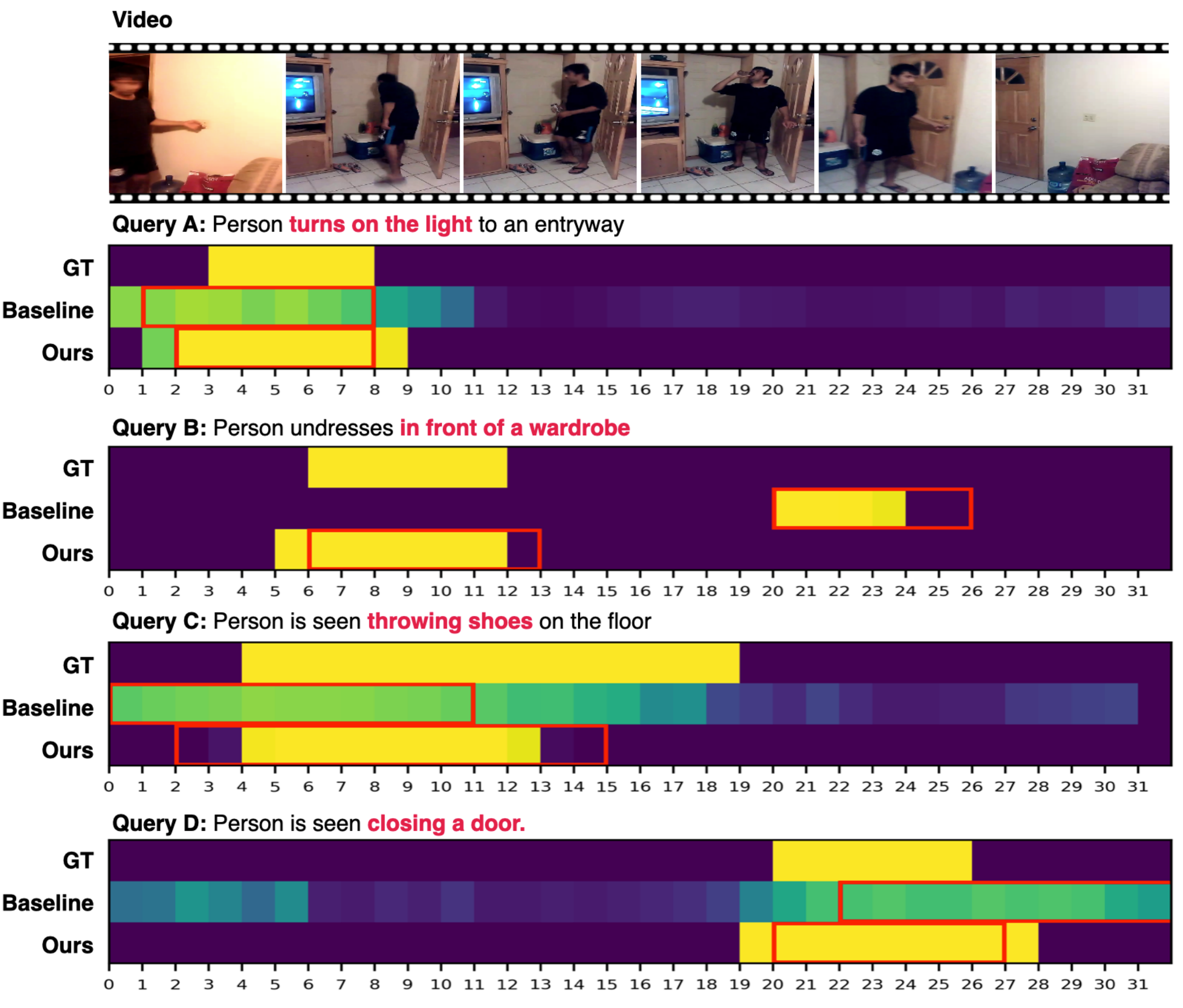}
  \end{center}
  \vspace{-4mm}
  \caption{
  Visualization of model's predictions. 
  We present the attention score~$\mathbf{o}$ (in Equation~\ref{eq: att}) and predicted moments (\textcolor{red}{red box}).
  }
  \label{fig:visualization1}
    \vspace{-3mm} 
\end{figure}

In Figure~\ref{fig:visualization1}, we visualize the predictions made by both the baseline and BM-DETR.
We can see that the baseline model is facing difficulties in accurately predicting video moments.
Although some of its predictions appear to be close to the ground-truth moments, its attention scores are dispersed rather than focused on the ground-truth moment.
On the other hand, BM-DETR demonstrates precise predictions with attention scores concentrated directly on the ground-truth moments.
Surprisingly, BM-DETR not only achieves accurate predictions but also, in some instances, refines them better than the ground-truth moment itself. 
For example, Query C doesn't explicitly mention ``drink'', but it's included in the latter part of the ground-truth moment.
Surprisingly, BM-DETR effectively identifies only the moment described in the query, demonstrating its effectiveness.

\section{Conclusion}
In this paper, we have argued that the inherent weak alignment problem in video datasets poses a significant hurdle to achieving successful VMR, underscoring the necessity for a robust model to address this challenge.
In response, we proposed the Background-aware Moment DEtection TRansformer (BM-DETR), a contrastive approach tailored for VMR, seamlessly incorporating background information into the moment prediction. 
Through extensive evaluations, we show that BM-DETR achieved remarkable improvements in performance and efficiency over the state-of-the-art methods across four widely-used VMR benchmarks.
We hope that our findings can contribute to future advancements in VMR.

\noindent\textbf{Limitations.}
While our model has been designed to perform robustly in the presence of noise and weak alignment in the training datasets, these issues persist because we have not directly addressed the quality of the datasets themselves.
As previously mentioned, approaches such as noise reduction or dataset re-annotation could help mitigate these problems, but they are impractical and fall outside the scope of our primary objective. A more feasible solution may involve generating clean and detailed video captions using language models~\cite{jung2022modal, zhao2023learning}.
In future work, we plan to explore these solutions and expand our method to a broader range of video understanding tasks.

\noindent\textbf{Acknowledgements.} This work was partly supported by the IITP (RS-2021-II212068-AIHub/10\%, RS-2021-II211343-GSAI/15\%, RS-2022-II220951-LBA/15\%, RS-2022-II220953-PICA/20\%), NRF (RS-2024-00353991-SPARC/20\%, RS-2023-00274280-HEI/10\%), and KEIT (RS-2024-00423940/10\%) grant funded by the Korean government.

%%%%%%%%% REFERENCES
{\small
\bibliographystyle{ieee_fullname}
\bibliography{egbib}
}
\newpage

\end{document}